\documentclass[10pt]{article}
\usepackage{amsfonts}
\usepackage{setspace}
\usepackage{amsmath}
\usepackage{amsthm}
\usepackage{amssymb}
\usepackage{bbm}
\usepackage{graphicx}
\usepackage[breaklinks,plainpages=false,colorlinks=true,linkcolor=blue,
  anchorcolor=green,citecolor=blue,filecolor=blue,urlcolor=blue]{hyperref}
\usepackage[round,authoryear]{natbib}
\usepackage[english]{babel}

\newcommand{\given}{\operatorname{|}}
\newcommand{\Pa}{\boldsymbol\Pi}
\newcommand{\Prob}{\operatorname{P}}

\title{Bayesian Network Structure Learning with Permutation Tests}

\author{Marco Scutari${}^*$ and Adriana Brogini \\ 
  Department of Statistical Sciences, University of Padova, Italy \\
  ${}^*$\textbf{Corresponding Author:} m.scutari@ucl.ac.uk}

\begin{document}

\maketitle

\begin{abstract}

  In literature there are several studies on the performance of Bayesian network
  structure learning algorithms. The focus of these studies is almost always
  the heuristics the learning algorithms are based on, i.e. the maximisation
  algorithms (in score-based algorithms) or the techniques for learning the
  dependencies of each variable (in constraint-based algorithms). In this paper
  we investigate how the use of permutation tests instead of parametric ones
  affects the performance of Bayesian network structure learning from discrete
  data. Shrinkage tests are also covered to provide a broad overview of the
  techniques developed in current literature.
  
  
\end{abstract}

\let\thefootnote\relax\footnotetext
  {This is a preprint of an article whose final and definitive form has been
  published in the Communications in Statistics -- Theory and Methods 
  (© 2012 Taylor \& Francis Group, LLC); Communications in Statistics -- Theory
  and Methods is available online at: www.tandfonline.com/10.1080/03610926.2011.593284.}
\thefootnote

\section{Introduction and Background}

Bayesian networks are a class of \textit{graphical models}, which allow an
intuitive representation of the probabilistic structure of multivariate data
using graphs \citep{pearl,koller}. They are composed by two parts:
\begin{itemize}
  \item a set of random variables $\mathbf{X} = \{X_1, X_2, \ldots, X_p\}$
    describing the features of the data. The probability distribution of
    $\mathbf{X}$ is called the \textit{global distribution} of the data, while
    the ones associated with each $X_i \in \mathbf{X}$ are called \textit{local
    distributions}. 
  \item a \textit{directed acyclic graph} (DAG), denoted $G = (\mathbf{V}, A)$.
    Each node $v \in \mathbf{V}$ is associated with one variable $X_i$, and
    they are often referred to interchangeably. The directed arcs $a \in A$
    that connect them represent direct stochastic dependencies; so if there is
    no arc connecting two nodes, the corresponding variables are either marginally
    independent or conditionally independent given a subset of the remaining
    variables.
\end{itemize}
In other words, each local distribution is associated with a single node $X_i$
and depends only on its parents (i.e. the nodes $X_j$ in $\mathbf{X}$ such
that $X_j \rightarrow X_i$, usually denoted by $\Pa_{X_i}$). This property,
which is known as the \textit{Markov property} of Bayesian networks \citep{pearl},
specifies the form of the decomposition of the global distribution into the
local ones:
\begin{equation}
  \Prob\left(\mathbf{X}\right) = \prod_{i=1}^p \Prob\left(X_i \given \Pa_{X_i}\right).
\end{equation}
In principle there are many possible choices for both the global and the local
distribution functions, depending on the nature of the data and the aims of the
analysis. However, literature have focused mostly on two cases: the \textit{discrete
case} \citep{heckerman}, in which both the global and the local distributions
are multinomial random variables, and the \textit{continuous case} \citep{heckerman3},
in which the global distribution is multivariate normal and the local distributions
are univariate normal random variables. In the first case, the parameters of
interest are the \textit{conditional probabilities} associated with each variable,
usually represented as \textit{conditional probability tables} (CPTs); in the
latter, the parameters of interest are the \textit{partial correlation coefficients}
between each variable and its parents.

The task of fitting a Bayesian network is called \textit{learning}, a term borrowed
from expert systems theory and artificial intelligence, and in general is implemented
in two steps.

The first step consists in finding the graph structure that encodes the conditional
independencies present in the data. Ideally it should coincide with the dependence
structure of the global distribution, or it should at least identify a distribution
as close as possible to the correct one in the probability space. This step is
called \textit{network structure} or simply \textit{structure learning} \citep{korb,
koller}, and is similar in approaches and terminology to model selection procedures
for classical statistical models. Several algorithms have been presented in
literature for this problem, thanks to the application of many results from
probability, information and optimisation theory. Despite the (sometimes confusing)
variety of theoretical backgrounds and terminology they can all be traced to only
three approaches: \textit{constraint-based} (which are based on conditional
independence tests), \textit{score-based} (which are based on goodness-of-fit
scores) and \textit{hybrid} (which combine the previous two approaches).

The second step is called \textit{parameter learning} and, as the name suggests,
deals with the estimation of the parameters of the global distribution. Assuming
the graph structure is known from the previous step, this can be done efficiently
by estimating the parameters of the local distributions.

In literature there are several studies on the performance of Bayesian network
structure learning algorithms; one of the most extensive performed in recent years
is presented in \citet{mmhc}. The focus of these studies is almost always the
heuristics the learning algorithms are based on, i.e. the maximisation algorithms
used in score-based algorithms or the techniques for learning the dependence
structure associated with each node in constraint-based algorithms. The influence
of the other components of the overall learning strategy, such as the conditional
independence tests (and the associated type I error threshold) or the network
scores (and the associated parameters, such as the equivalent sample size), is
usually not investigated. However, limiting such studies to the performance of
heuristics poses serious doubts on their conclusions, because the 
decisions made by the heuristics are based on the values of the statistical criteria
they use to extract information from the data. Therefore, it is important to
choose a conditional independence test or a network score presenting a good
behaviour for the data at hand and to tune it appropriately. 

For this reason, in this paper we will investigate the behaviour of permutation
conditional independence tests and tests based on shrinkage estimators for discrete
data. These two classes of tests are usually not considered in literature, where
the asymptotic $\chi^2$ tests based on Pearson's $\mathrm{X}^2$ \citep{fienberg}
and mutual information \citep{kullback} are the \textit{de facto} standard.
In particular, we will study the permutation Pearson's $\mathrm{X}^2$ test and
the permutation mutual information test described in \citet{edwards}, and the
shrinkage test based on the estimator for the mutual information presented in
\citet{shrinkent}.

\section{Conditional Independence Tests}

We will now introduce the conditional independence tests whose performance will
be considered in Section \ref{sec:traditional}. Since we are limiting ourselves
to discrete data, both the global and the local distributions are assumed to be
multinomial, and the latter are represented as conditional probability tables.
Conditional independence tests and network scores for discrete data are functions
of these conditional probability tables through the observed frequencies
$\{n_{ijk}, i = 1, \ldots, R, j = 1, \ldots, C, k = 1, \ldots, L\}$ for the
random variables $X$ and $Y$ and all the configurations of the levels of the
conditioning variables $\mathbf{Z}$.

\subsection{Parametric Tests}

Two classic conditional independence tests used in the analysis of contingency
and probability tables are:
\begin{itemize}
  \item \emph{mutual information}: an information-theoretic distance measure
    defined as
    \begin{equation}
      \mathrm{MI}(X, Y \given \mathbf{Z}) = \sum_{i=1}^R \sum_{j=1}^C \sum_{k=1}^L
        \frac{n_{ijk}}{n} \log\frac{n_{ijk} n_{++k}}{n_{i+k} n_{+jk}}.
    \end{equation}
    It is proportional to the log-likelihood ratio test $\mathrm{G}^2$ (they
    differ by a $2n$ factor, where $n$ is the sample size) and is related to
    the deviance of the tested models.

  \item \emph{Pearson's $X^2$}: Pearson's $\mathrm{X}^2$ test for contingency
    tables,
    \begin{align}
      &\mathrm{X}^2(X, Y \given \mathbf{Z}) = \sum_{i=1}^R \sum_{j=1}^C \sum_{k=1}^L
        \frac{\left(n_{ijk} - m_{ijk}\right)^2}{m_{ijk}},&
      &\text{where}&
      &m_{ijk} = \frac{n_{i+k}n_{+jk}}{n_{++k}}.
    \end{align}
\end{itemize}
The asymptotic null distribution is $\chi^2$ with $(R - 1)(C - 1)L$ degrees of
freedom in both cases. For a detailed analysis of their properties we refer
the reader to \citet{agresti} and \citet{fienberg}. The main limitation of these
tests is the rate of convergence to their limiting distribution, which is
particularly problematic when dealing with small samples and sparse contingency
tables. This situation, which is often referred to as ``small $n$, large $p$'',
is very common in many settings in which Bayesian networks are used (such as
gene expression and omics data). 

\subsection{Permutation Tests}

The mutual information and Pearson's $\mathrm{X}^2$ tests can also be performed
by conditioning on a sufficient statistic and using the permutation distribution
as the null distribution \citep{pesarin}. The observed significance values are
then computed with a \textit{conditional Monte Carlo} (CMC) simulation, as detailed
in \citet{edwards}:
\begin{enumerate}
  \item compute the value of the test statistic for the original data set,
    and denote it with $T$;
  \item perform the following steps for a suitable number $R$ of times, usually
    between $500$ and $5000$:
    \begin{enumerate}
      \item randomly permute the observations presenting the same configuration
        of the conditioning variables $\mathbf{Z}$; this is typically done by
        applying the permutation algorithm from \citet{patel} to the
        contingency tables associated with the configurations of $\mathbf{Z}$;
     \item compute the test statistic on the resulting data, and denote it
        with $T^*_r$, $r = 1, \ldots, R$;
    \end{enumerate}
  \item compute the significance value of $T$ as
  \begin{equation}
    \hat\alpha_R = \frac{1}{R} \sum_{r = 1}^R \mathbbm{1}_{\{x \geqslant T\}}(T^*_r).
  \end{equation}
\end{enumerate}
Both the permutation algorithm by \citet{patel} and the conditional independence
tests considered in this paper have the marginal totals $\{n_{i+k}\}$ and
$\{n_{+jk}\}$ as sufficient statistics, so they can be computed efficiently.

The main advantage of permutations tests is that they do not require a large sample
size or particular distributional assumptions to perform well, because they operate
conditioning on the available data \citep{pesarin}. Therefore, they perform better
than the parametric tests usually found in literature, because they are not
limited by the rate of convergence to the respective asymptotic distributions.
However, the computer time required by the generation of the permutations of the
data and by the repeated evaluation of the test statistic have prevented their
widespread use in many settings in which high-dimensional problems are the
norm.

\subsection{Shrinkage Tests}

In high-dimensional, multivariate problems, the maximum likelihood estimator
is known to be inefficient and displays a considerable instability for most
reasonable, finite sample sizes. This phenomenon, which is known as the
``curse of dimensionality'', is caused by the exponential increase in the
number of parameters as the number of variables increase.

These issues can be explained as a consequence of the inadmissibility of the
maximum likelihood estimator for the mean of multivariate distributions
discovered by \citet{stein} and investigated by \citet{jstein}. A solution is
provided in the form of a \textit{regularised estimator}, which includes some
bias in order to increase the overall performance of the estimator. Since the
natural parameters of the test statistics we are considering are the probabilities
$\{p_{ijk}\}$ associated with the observed frequencies $\{n_{ijk}\}$, we will
denote such an estimator as $\mathbf{\tilde{p}} = \{\tilde{p}_{ijk}\}$ and
the maximum likelihood estimator as $\mathbf{\hat{p}} = \{\hat{p}_{ijk} = 
n_{ijk} / n\}$. The regularised estimator is then defined as a linear combination
of the maximum likelihood estimator and a \textit{target distribution} with
probabilities $\mathbf{t} = \{t_{ijk}\}$, which is usually chosen to be
uniform (i.e. $t_{ijk} = 1 / |\mathbf{t}|$):
\begin{align}
  &\tilde{p}_{ijk} = \lambda t_{ijk} + (1 - \lambda) \hat{p}_{ijk},& &\lambda \in [0,1].
\end{align}
Such an estimator is called a \textit{shrinkage estimator}, because $\mathbf{\hat{p}}$
is shrunk towards $\mathbf{t}$ in the parameter space; $\lambda$ is likewise
called the \textit{shrinkage coefficient}.

A closed-form estimator for $\lambda$ has been derived by \citet{ledoit} as
the value that minimises the mean squared error of $\mathbf{\tilde{p}}$.
\citet{shrinkent} derived its expression for multinomial probabilities, with the
aim of defining an improved entropy estimator \citep{kullback}; it has the form
\begin{equation}
  \lambda^* = \frac{1 - \sum \hat{p}_{ijk}^2 }{ (n - 1) \sum (t_{ijk} - \hat{p}_{ijk})^2}.
\end{equation}
The application of this result to the mutual information test leads to the
definition of the corresponding \textit{shrinkage mutual information test},
which is based on the shrinkage estimator $\mathbf{\tilde{p}}$ instead of
the usual maximum likelihood estimator $\mathbf{\hat{p}}$. It has the form
\begin{equation}
  \mathrm{MI}_{sh}(X, Y \given \mathbf{Z}) = \sum_{i=1}^R \sum_{j=1}^C \sum_{k=1}^L
    \tilde{p}_{ijk} \log \frac{ \tilde{p}_{ijk} \tilde{p}_{++k} }{ \tilde{p}_{i+k} \tilde{p}_{+jk} }.
\end{equation}
The observed significance value can still be computed using the asymptotic
$\chi^2$ distribution used for the classic mutual information test, for two
reasons. First, \citet{ledoit} proved that $\lambda^*$ converges to zero as the
sample size diverges, which means that the shrinkage test $\mathrm{MI}_{sh}$
and the classic parametric test $\mathrm{MI}$ have the same asymptotic behaviour.
Furthermore, the shrinkage mutual information test is still a log-likelihood ratio
test. The only difference is in the number of the parameters, as both the null
and the observed distributions gain an additional parameter, $\lambda$. Therefore,
according to \citet{lehmann} the asymptotic distribution of the shrinkage test
is again $\chi^2$ with $(R - 1)(C - 1)L$ degrees of freedom.

\section{Effects on Structure Learning}
\label{sec:traditional}

We will now investigate the behaviour of the permutation conditional independence
tests and shrinkage tests introduced in the previous section, using the parametric
tests as a reference. Five performance indicators will be taken into consideration:
\begin{itemize}
  \item the posterior density of the network for the data it was learned from,
    as a measure of goodness of fit. It is known as the \textit{Bayesian Dirichlet
    equivalent} score (BDe) from \citet{heckerman} and has a single parameter, the
    \textit{equivalent sample size}, which can be thought of as the size of an 
    imaginary sample supporting the prior distribution. The equivalent sample size
    will be set to $10$ as suggested in \citet{koller};
  \item the BIC score \citep{bic} of the network for the data it was learned from,
    again as a measure of goodness of fit;
  \item the posterior density of the network for a new data set, as a measure of
    how well the network generalises to new data;
  \item the BIC score of the network for a new data set, again as a measure of
    how well the network generalises to new data;
  \item the Structural Hamming Distance (SHD) between the learned and the true
    structure of the network, as a measure of the quality of the learned
    dependence structure \citep{mmhc}.
\end{itemize}
These indicators will be estimated for each test, using the bnlearn R
package \citep{R,jss09} as follows:
\begin{enumerate}
  \item generate a sample from the true probability distribution of the ALARM
    network from \citet{alarm}. ALARM contains $37$ nodes and $46$ arcs, for a total
    of $509$ parameters, and is frequently used as a benchmark in the literature of
    Bayesian networks \citep{larranaga,reinsertion,decampos};
  \item learn a network structure with the \mbox{Max-Min} \mbox{Hill-Climbing}
    (MMHC) hybrid algorithm \citep{mmhc} using one of the conditional independence
    tests under investigation and the BDe score. This learning strategy has been
    shown to be one of the most effective up to date; it combines the Max-Min Parents
    and Children (MMPC) constraint-based algorithm with a score-based hill climbing
    search. Two thresholds are considered for the type I error of the tests: $0.05$
    and $0.01$. Since results are very similar, they are reported only for
    $0.05$ for brevity;
  \item learn a second network structure from the same data with the
    asymptotic, parametric test based either on Pearson's $\mathrm{X}^2$ or on the
    maximum likelihood estimator for the mutual information, depending on which test
    was used in the previous step;
  \item repeat the previous two steps using the BIC score instead of BDe;
  \item compute the relevant performance indicators for each pair of network
    structures, and the differences are standardised to express the relative
    difference over the values obtained with the asymptotic tests. In particular,
    BDe will be only considered for networks learned in step 2 and BIC for networks
    learned in step 4.
\end{enumerate}
These steps will be repeated $50$ times for each sample size. The data set needed
to assess how well the network generalises to new data is generated again from the
true probability structure of the ALARM network and contains $20000$ observations.
The parameters of the network, which are the elements of the conditional probability
tables associated with the nodes of the networks, are estimated using the
corresponding empirical frequencies.

\subsection{Permutation Tests}

Nonparametric conditional independence tests, and permutation tests in particular,
provide a feasible alternative to the corresponding parametric tests in a wide range
of situations. 

\begin{figure}[p]
  \begin{center}
    \includegraphics[width=0.9\textwidth]{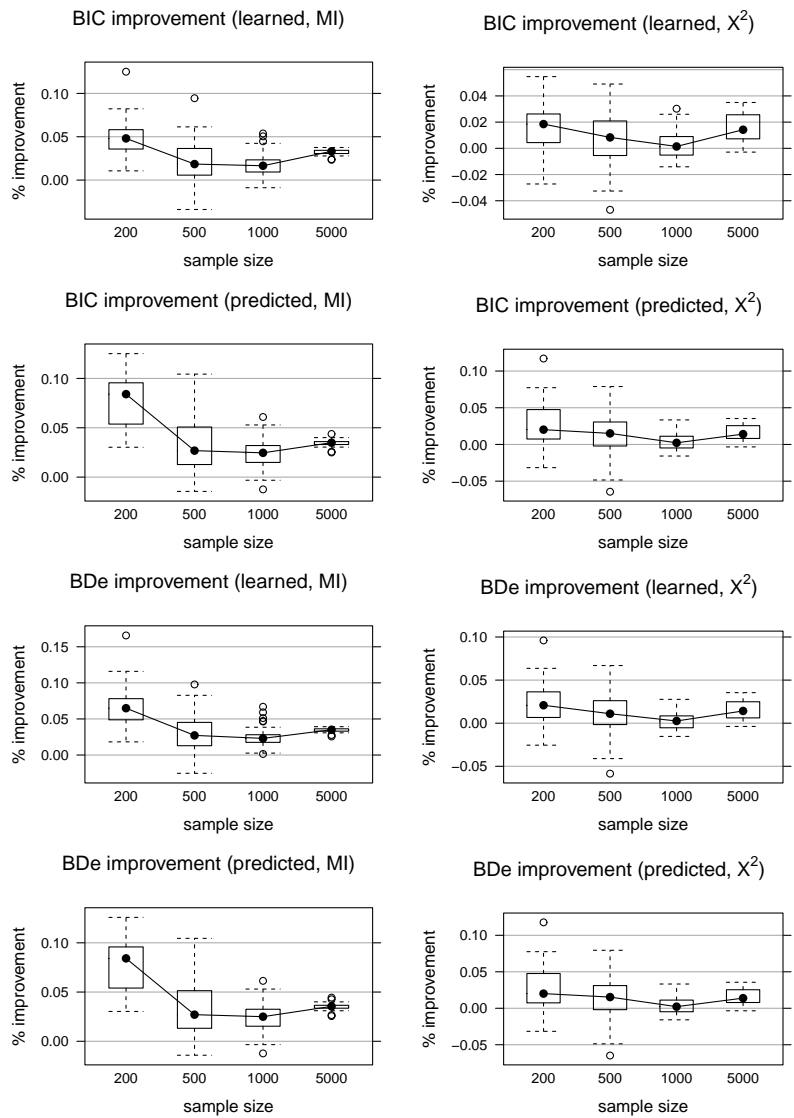}
  \end{center}
  \caption{Improvements in Bayesian network structure learning when using the
    permutation mutual information (on the left) and the permutation Pearson's 
    $\mathrm{X}^2$ (on the right) tests. The black dot in each box-plot
    represents the median.}
\label{fig:permtest}
\end{figure}

\begin{figure}[t]
  \begin{center}
    \includegraphics[width=0.9\textwidth]{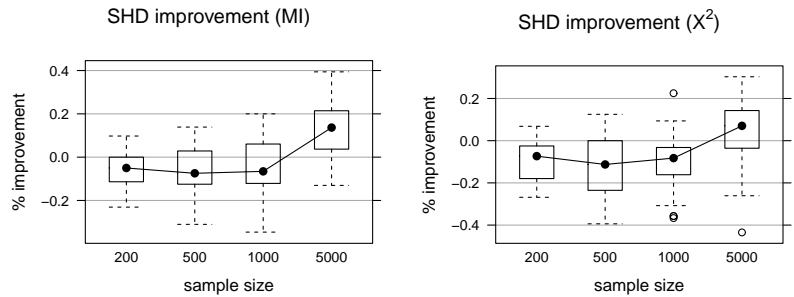}
  \end{center}
  \caption{Differences in the Structural Hamming Distance when using the permutation
    mutual information (on the left) and the permutation Pearson's $\mathrm{X}^2$
    (on the right) tests. The black dot in each box-plot represents the median.}
\label{fig:permshd}
\end{figure}

The effects of the properties of the permutation Pearson's $\mathrm{X}^2$ and the
permutation mutual information tests on Bayesian network structure learning are
shown in Figure \ref{fig:permtest} and Figure \ref{fig:permshd}. First, we can
clearly see from the box-plots in Figure \ref{fig:permtest} that the use of
permutation tests results in network structures with higher scores for all the
considered sample sizes ($200$, $500$, $1000$ and $5000$). This is also true when
considering the new data set, meaning that the network structures learned with
these tests are better for predicting the behaviour of new samples. As expected,
improvements in the BIC and BDe scores are particularly significant for low
sample sizes; the probability structure of the ALARM network has $509$ parameters,
which means that the ratios between the number of observations and the number of
parameters are $0.3929$, $0.9823$, $1.9646$ and $9.8231$ respectively.

It is also interesting to note that, even though the performance of parametric
tests improves with the sample size, both permutation tests appear to improve at
a faster rate. In fact, in all plots in Figure \ref{fig:permtest} the relative
improvement for samples of size $5000$ is greater than the corresponding
improvement for samples of size $2000$, regardless of the score we are considering
or the data set it is computed from.

On the other hand, the network structures learned with permutation
tests considered in this section are often not as close to the true network
structure as the ones learned with the corresponding parametric tests. This
is can be clearly seen from the box-plots in Figure \ref{fig:permshd}, which
show that in the majority of simulations the relative difference between the
SHD values is negative (i.e. the SHD associated with the parametric test
is smaller than the SHD associated with the permutation test). Permutation tests
outperform parametric tests only for samples of size $5000$.

The comparatively poor performance of permutation tests in terms of SHD can be
attributed to the conditioning on the observed sample that characterises them.
Most of the samples considered in this analysis are too small to provide an
adequate representation of the true probability structure of the ALARM network,
as evidenced by the ratios between their sample sizes and the number of
parameters. Therefore, the network structures learned with permutation tests
from these samples are able to capture only part of the true dependence
structure. The arcs that are most likely to be missed, however, are those
that represent the weakest dependence relationships; otherwise the networks
would not be able to fit new data so well.

In conclusion, permutation tests result in better network structures than the
corresponding parametric tests, both in terms of goodness of fit and in how
well the networks are able to generalise to new data. However, if the focus
of the analysis is the structure of the network itself (such as when the
Bayesian network is considered as a causal model) parametric tests may be
preferable for small samples.

\begin{figure}[p]
  \begin{center}
    \includegraphics[width=0.9\textwidth]{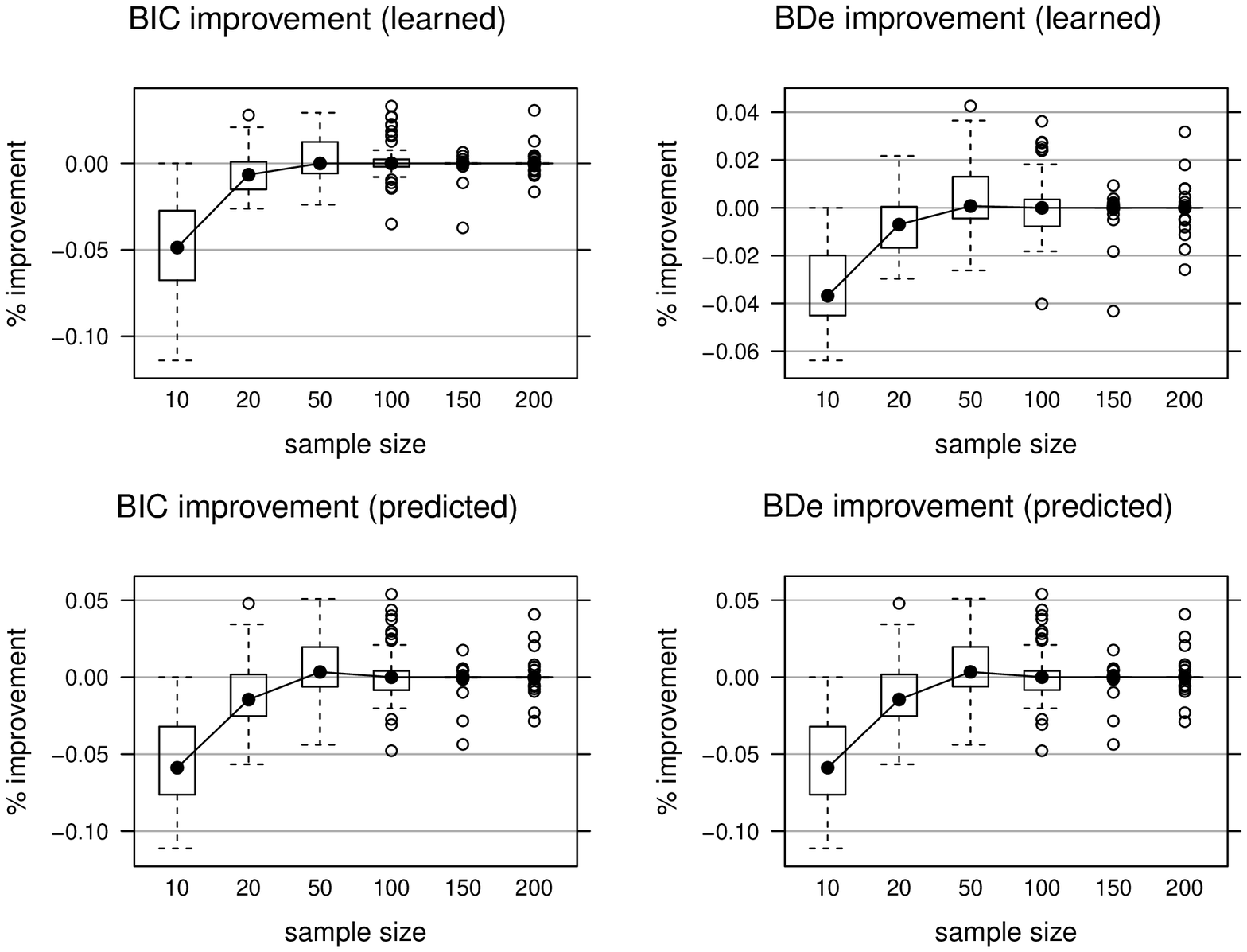}
  \end{center}
  \caption{Improvements in Bayesian network structure learning when using the
    shrinkage estimator for the mutual information. The black dot in each box-plot
    represents the median.}
\label{fig:shtest}
  \begin{center}
    \includegraphics[width=0.5\textwidth]{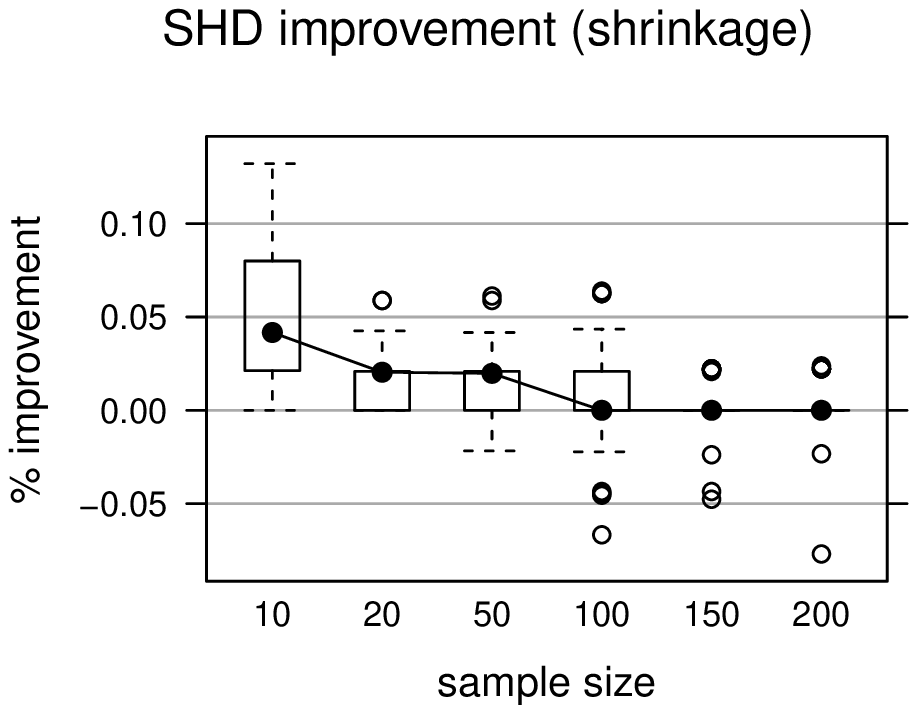}
  \end{center}
  \caption{Differences in the Structural Hamming Distance when using the shrinkage
    estimator for the mutual information. The black dot in each box-plot represents
    the median.}
\label{fig:shshd}
\end{figure}

\subsection{Shrinkage Tests}
\label{sec:shrinkeval}

The shrinkage mutual information test has a completely different behaviour
than the permutation tests covered above.

As expected from a test based on a regularised estimator, the networks learned
using shrinkage tests do not fit the data as well as the networks learned
with the corresponding maximum likelihood tests. This can be clearly seen from
the box-plots in Figure \ref{fig:shtest}. The relative differences in the BIC and
BDe scores are almost never positive for either the data the networks have been
learned from or the new data, in particular for samples of size $10$ and $20$.
Such small samples are most likely to result in sparse contingency tables, and
therefore in high values of the shrinkage coefficient, as soon as a few
conditioning variables are included in the tests. Larger samples are less
affected by the regularisation of the shrinkage estimator, because the shrinkage
coefficient converges to zero as the number of observations diverges
\citep{ledoit}. This means that for larger samples (i.e. $100$, $150$ and $200$)
the behaviour of the shrinkage mutual information test approaches the one of
the classic mutual information test, as can be seen from the increasingly
small difference between the two in terms of BIC and BDe scores.

An important side effect of the regularisation performed by the shrinkage
estimator is the reduction of the structural distance from the true network
structure for small samples. We can see from Figure \ref{fig:shshd} that the
shrinkage test outperforms the test based on the maximum likelihood estimator;
there is a systematic improvement for sample sizes $10$, $20$ and $50$ (i.e.
SHD is smaller for the shrinkage test). As the sample size increases, the
behaviour of the shrinkage test approaches again the one of the corresponding
maximum likelihood test. These simulations confirm the results produced with
shrinkage tests for many ``small $n$, large $p$'' problems, such as those
studied in \citet{ss05} and \citet{nicole}, which have led to a widespread
use of shrinkage tests in biology and genetics.

\section{Conclusions}

In this paper we investigated how the use of permutation tests instead of 
parametric ones affects the performance of Bayesian network structure learning
from discrete data, while also covering shrinkage tests. Permutation tests
result in better network structures than the corresponding parametric tests,
both in terms of goodness of fit and in how well the networks are able to 
generalise to new data. Shrinkage tests, on the other hand, outperform both
parametric and permutation tests in the quality of the network structure itself,
which is closer to the true dependence structure of the data.

\end{document}